\documentclass{iosart2c}

\usepackage[T1]{fontenc}
\usepackage{times}%

\usepackage{natbib}
\usepackage{dcolumn}
\usepackage{amsmath,amssymb,amsfonts}
\usepackage{algorithmic}
\usepackage{graphicx}
\usepackage{textcomp}
\usepackage{multirow}
\usepackage{float}
\usepackage{booktabs}
\usepackage{makecell}
\usepackage{bbding}
\usepackage{url}

\setcitestyle{numbers,square}

\newcolumntype{d}[1]{D{.}{.}{#1}}

\firstpage{1} \lastpage{5} \volume{1} \pubyear{2009}

\begin{document}
\begin{frontmatter}                           

%
\title{MNeRV: A Multilayer Neural Representation for Videos}

\runningtitle{MNeRV: A Multilayer Neural Representation for Videos}

\author[A]{\fnms{Qingling} \snm{Chang}},
\author[A]{\fnms{Haohui} \snm{Yu}},
\author[B]{\fnms{Shuxuan} \snm{Fu}},
\author[A]{\fnms{Zhiqiang} \snm{Zeng}},
and
\author[A]{\fnms{Chuangquan} \snm{Chen}}
\address[A]{The Institute of China-Germany Artificial Intelligence, Wuyi university, Welcome Avenue, 99, Jiangmen, China
}
\address[B]{College of Mathematics and Physics, North China Electric Power University, No.2 Beinong Road, Huilongguan, Changping District, Beijing, China
}
E-mail: cqlxst@126.com.com\\
E-mail: Yhh17666499405@163.com\\
E-mail: IantheFu@ncepu.edu.cn\\
E-mail: zhiqiang.zeng@outlook.com\\
E-mail: chenchuangquan87@163.com\\

\begin{abstract}
As a novel video representation method, Neural Representations for Videos (NeRV) has shown great potential in the fields of video compression, video restoration, and video interpolation. In the process of representing videos using NeRV, each frame corresponds to an embedding, which is then reconstructed into a video frame sequence after passing through a small number of decoding layers (E-NeRV, HNeRV, etc.). However, this small number of decoding layers can easily lead to the problem of redundant model parameters due to the large proportion of parameters in a single decoding layer, which greatly restricts the video regression ability of neural network models. In this paper, we propose a multilayer neural representation for videos (MNeRV) and design a new decoder M-Decoder and its matching encoder M-Encoder. MNeRV has more encoding and decoding layers, which effectively alleviates the problem of redundant model parameters caused by too few layers. In addition, we design MNeRV blocks to perform more uniform and effective parameter allocation between decoding layers. In the field of video regression reconstruction, we achieve better reconstruction quality (+4.06 PSNR) with fewer parameters. Finally, we showcase MNeRV’s performance in downstream tasks such as video restoration and video interpolation. The source code of MNeRV is available at https://github.com/Aaronbtb/MNeRV.
\end{abstract}

\begin{keyword}
Neural networks\sep Noise reduction\sep Video compression\sep Network architecture\sep Decoding
\end{keyword}

\end{frontmatter}

\section{Introduction}
\label{sec:introduction}
\quad With the rise of short videos on the internet, more and more people are uploading videos shot on mobile devices (such as smartphones) to video-sharing websites like YouTube and Bilibili. Videos are playing an increasingly important role in people’s lives. Due to objective factors such as network bandwidth and storage space, videos must be compressed to better serve people. The growing demand for network video sharing and the limitations of bandwidth and storage make video compression algorithms more important. Traditional video compression methods\cite{wiegand_overview_2003,sullivan_overview_2012,bross_overview_2021} can explicitly represent videos as frame sequences, but they bring huge computational costs during compression and decompression, and the corresponding decoding time cost also increases. Some researchers have introduced deep learning into video encoding and decoding. Based on the traditional framework, some modules are replaced with trainable deep learning models\cite{afonso_video_2019,wang_multi-scale_2020,yan_convolutional_2019,mao_convolutional_2020,yang_deep_2020,ma_convolutional_2020,zhang_recursive_2020,noauthor_high-definition_nodate}, which has greatly improved the performance. Other researchers have replaced all traditional modules with deep learning models\cite{haojie_liu_learned_nodate,hu_coarse--fine_2020,balle_end--end_2016,johnston_improved_2018,kim_adversarial_2020,noauthor_nonlinear_nodate,ishikawa_feedback_2021,goodfellow_generative_2014,ignatov_ai_2019} on the basis of the traditional framework, and adopted an end-to-end training method, which has also achieved good compression results. However, these two methods have similar disadvantages to traditional methods: high training computational cost and slow decoding speed. To avoid the high computational cost of the traditional pipeline and explore the more powerful fitting ability of neural networks, some researchers have applied implicit neural representations to video encoding and decoding, proposing a new video representation method: neural representations for videos (NeRV)\cite{chen_nerv_2021}. This method represents videos as neural networks in an implicit neural representation way at the decoder end.It takes video frames as indices, inputs them into the decoding layer for decoding, and outputs images, that is, representing videos as an implicit neural network. This unique method {NeRV and other series of papers} has not only shown good results in the compression field, but also achieved good results in various video downstream task such as video interpolation and video denoising, with the advantages of easy training and fast decoding. However, most of the models in the NeRV-like for video representation have relatively few decoding layers in the decoding end, which leads to two problems. The first problem is that fewer decoding layers perform poorly in fitting videos with large camera movements. The second problem is that fewer NeRV-like blocks lack the ability to allocate model parameters reasonably.

In this work, we propose a multilayer neural representation for videos (MNeRV) that has a new encoder end: M-Encoder and decoder end: M-Decoder. Thanks to the design of more encoding and decoding layers, it has excellent performance in fitting videos with large camera movements. Additionally, we propose an MNeRV block that allows for more reasonable parameter allocation at the decoder end. We also removed some redundant designs, making the model more streamlined and able to achieve better results with fewer parameters. In M-Encoder, we keep the number of encoding layers the same as the number of decoding layers in the decoder end, so that the extracted feature indices match the decoder end more closely. We also introduce the Global Response Normalization (GRN)\cite{10205236} layer to enhance the competition of feature channels at the encoding end and improve the model’s accuracy. We apply MNeRV to downstream tasks such as video compression, video restoration, and video interpolation, and demonstrate its excellent performance.

Our main contributions are summarized as follows:

\begin{itemize}
	\item[$\bullet$] We propose MNeRV, a novel image-wise video implicit representation method with efficient multilayer.
	\item[$\bullet$] We design a new encoder and decoder: M-Encoder, M-Decoder. By introducing the GRN layer, removing redundant structures, and increasing the number of encoding and decoding layers, the model is more efficient. We design ablation experiments to prove that this is effective.
	\item[$\bullet$] Through extensive experiments, we demonstrate that our method has better performance (+4.06 PSNR) in video reconstruction quality and better results in downstream tasks such as video compression, video restoration, and video interpolation.
\end{itemize}

\section{Related Work}
\subsection{Pixel-wise Implicit Neural Representations}
\quad Implicit neural representations are a novel signal representation method that approximates a mapping function by fitting a neural network. Implicit neural representations have powerful modeling capabilities for various signals, such as data compression\cite{dupont_coin_2021,dupont_coin_2022,zhang_implicit_2022}, 3D reconstruction\cite{li_neural_2021,littwin_deep_2019,niemeyer_differentiable_2020,wang_neus_2021,xian_space-time_2021}, and 3D view synthesis\cite{li_neural_2021,littwin_deep_2019,niemeyer_differentiable_2020,wang_neus_2021,xian_space-time_2021}, etc. In the early days of implicit neural representations for videos, it was usually pixel-wise encoding and decoding, which specifically trained the model to learn the mapping relationship between the coordinates of a point and its RGB value\cite{bauer_spatial_2023,kim_scalable_2022}. This method has high training costs, slow encoding and decoding speeds, and lower compression rates. 
\subsection{Image-wise Implicit Neural Representations}
NeRV first proposed an implicit representation method for image-wise videos, which trains and fits videos into a neural network using convolution and pixel shuffle. In this way, the decoding process of the video is transformed into the inference process of the model, which greatly improves the decoding speed of the video. In addition, due to the characteristics of neural networks themselves, NeRV-like videos also perform well in downstream video tasks such as video interpolation and video restoration, attracting more and more researchers to study and improve it. Bai et al\cite{bai_ps-nerv_2022} balanced the coordinate-based implicit neural representation (INRs) and the image-based implicit neural representation (NeRV). They introduced the idea of partitioning into NeRV, representing videos as multiple image blocks, each with a coordinate, and adding AdaIN to the network to improve the fitting effect. Li et al\cite{li_e-nerv_2022} improved the encoder and decoder layers based on NeRV and proposed E-NeRV, further removing redundant structures in the neural network and introducing a spatiotemporal context-based encoder, achieving an experimental effect that converges 8 times faster than NeRV. Mai et al\cite{9879067} introduced a motion-adjustable neural implicit video representation, which maps time to a driving signal to modulate the frame-generation process, and achieved good results. Lee et al\cite{lee_ffnerv_2022} further improved NeRV by introducing optical flow into the frame information. In addition, they introduced a fully convolutional architecture, enabled by one-dimensional temporal grids, improving the continuity of spatial features.

However, all of the above improvements are based on the positional embedding of time as the input. Time-based encoding cannot capture specific content information in the image, resulting in a low compression rate of the model. To make the embedding content-related, Chen et al. proposed a content-adaptive encoder CNeRV\cite{chen_cnerv_2022}. Subsequently, following the idea of feature embedding decoder, they proposed another decoder architecture HNeRV\cite{chen_hnerv_2023}, which uses the convnext block in the encoder to encode a smaller feature map of image as an embedding. In this way, the encoded embedding is content-related, resulting in a high compression rate and good fitting effect of the model. However, due to the fewer decoding layers at the decoder end, the accuracy will decrease when fitting some fast-moving objects, moving cameras, and other dynamic videos. He et al. fully utilized the fitting role of neural networks in various types of videos and proposed D-NeRV\cite{he_towards_nodate}, which uses massive video data to represent a large and diverse set of videos as a single neural network, employing the task-oriented flow as intermediate output to reduce spatial redundancies. It performs better in long videos and is one of the development directions of future large video models. However, the performance is poor when fitting a single short video. Zhao et al\cite{zhao_dnerv_2023} fully utilized the diff flow of frames on the basis of HNeRV and proposed a differential encoder to model the spatial features of specific content in a short time, achieving good results in video interpolation and video restoration. However, the model cannot maintain small embeddings, resulting in a low overall compression rate of the video represented by neural networks, which is a disaster for multi-frame videos. Kwan et al. proposed HiNeRV\cite{kwan_hinerv_2023}, which pursues the ultimate bit rate performance and is currently the most competitive INRs method in video compression. However, its decoding method based on bilinear interpolation performs poorly in video interpolation and video restoration.

In MNeRV, a single video is fitted with a neural network while maintaining small embeddings. Thanks to the reasonable parameter volume and streamlined network architecture at each layer, MNeRV can reconstruct videos with better quality using fewer parameters. Unlike DNeRV and HiNeRV, which focus on a specific downstream task of video, MNeRV is consistent with HNeRV in achieving good performance on many downstream tasks such as video interpolation, video restoration, and video compression.

\section{Preliminaries}
\begin{figure}
	\centering
	\includegraphics[width=3in,height=9in,keepaspectratio]{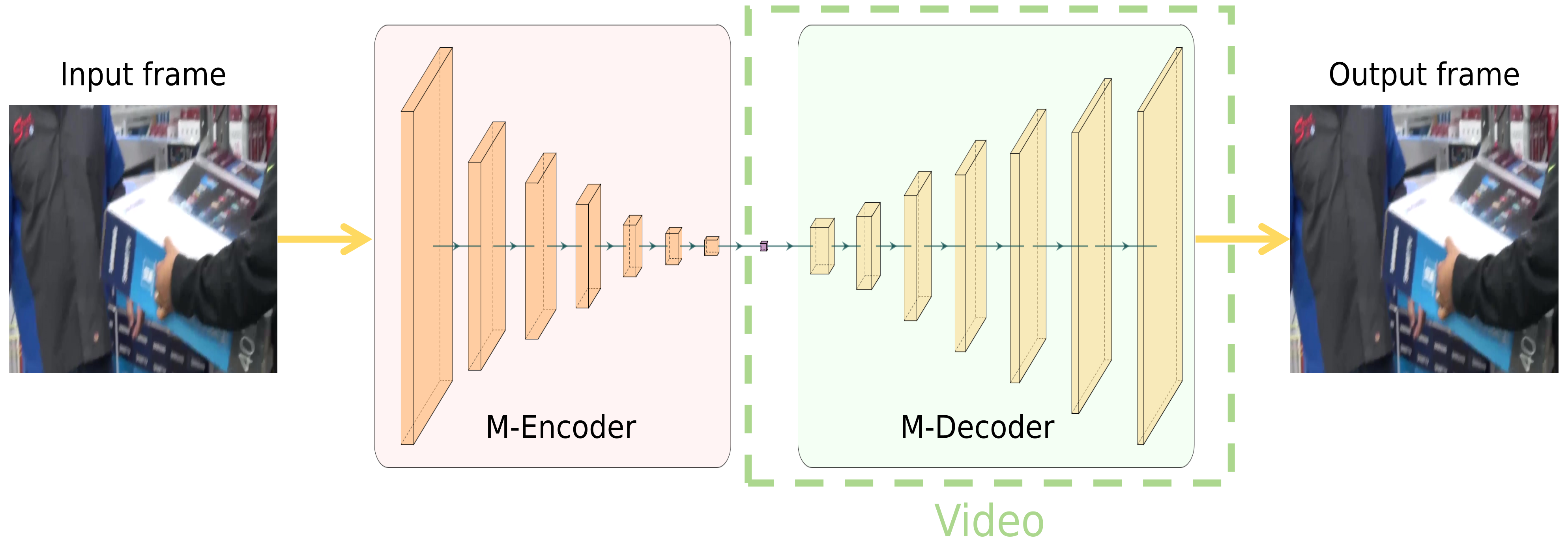}
	\caption{Architecture of multilayer neural representation for videos.}
	\label{architecture}
\end{figure}

\quad In NeRV-like models, the model mainly learns a mapping between the current frame $vt$ and the reconstructed frame $vt'$ after passing through the neural network, where $vt \in \mathbb{R}^{3\text{$\times$}H\text{$\times$}W}$. The entire architecture(See Figure \ref{architecture}) is divided into the encoder part $Fe$, the embedding $ft$, and the decoder part $Fd$, expressed as follows:
\begin{equation}
\begin{aligned}
&ft=Fe(vt)\\
&vt'=Fd(ft)
\end{aligned}
\end{equation}
$Fe$ is a learnable network encoder. In NeRV and E-NeRV, $Fe$ uses regular frequency positional encoding. In HNeRV, $Fe$ uses ConvNeXt blocks\cite{9879745} to construct. $ft$ is the feature map encoded by the encoder $Fe$. $Fd$ is the decoder with many NeRV-like blocks. NeRV-like blocks consist of three layers: Convolution layer, PixelShuffle layer, and Activation layer, where the activation layer and PixelShuffle layer do not contain learning parameters. Unlike the past coordinate-based pixel-wise implicit neural representation\cite{kim_scalable_2022}, NeRV blocks mainly learn the mapping from feature map to feature map. During the inference process of the model, the number of channels in the feature map will decrease after each NeRV block, while the size of the feature map will increase. Finally, the number of channels in the feature map will decrease to 3, and the size will increase to the size of the image. This is an image-wise representation method, which gives NeRV more compact model parameters and faster decoding speed. In the NeRV-like architecture before HNeRV, the positional embedding of time was used as input, and the embedding did not contain content information, resulting in a low model compression rate. HNeRV encodes a frame image into a small feature map and inputs it to the decoder as an embedding. Based on the content-based encoder end, it can quickly encode the content of a frame into a feature map, which eliminates the feature encoding process after embedding in NeRV and has the advantage of fast decoding compared to NeRV.

MNeRV and HNeRV encode $ft$ into entropy coding and input it into the decoder $Fd$ to obtain the reconstructed frame Vt’. As a type of implicit neural network representation, when MNeRV and HNeRV are used to represent videos, the total size includes both the embedding and decoder parts. Its principle is to fit several frames in a video into a neural network stored in the decoder end, and the size of the neural network directly affects the size of the fitted video. Therefore, the size of the video represented as a neural network is the sum of the parameters of the decoder $Fd$ and the embedding $ft$, expressed as follows:

\begin{equation}model size=ft+Fd\label{eq}\end{equation}

There are multiple decoding layers in the decoder $Fd$. In HNeRV, most decoding layers contain a NeRV block. MNeRV further removes redundant structures, changes the decoding step size, and changes the kernel size of the convolution layer in each MNeRV block, so that all decoding layers contain a MNeRV block, reducing the model parameters under the same effect (see SEC3.1 for details). In M-Encoder, we use ConvNeXt blocks to construct our encoder, introduce the GRN layer, and change the kernel size and step size (see SEC4.1 for details).

\section{Method}
\begin{figure*}
	\centering
	\includegraphics[width=150mm,trim=20 0 70 0,clip]{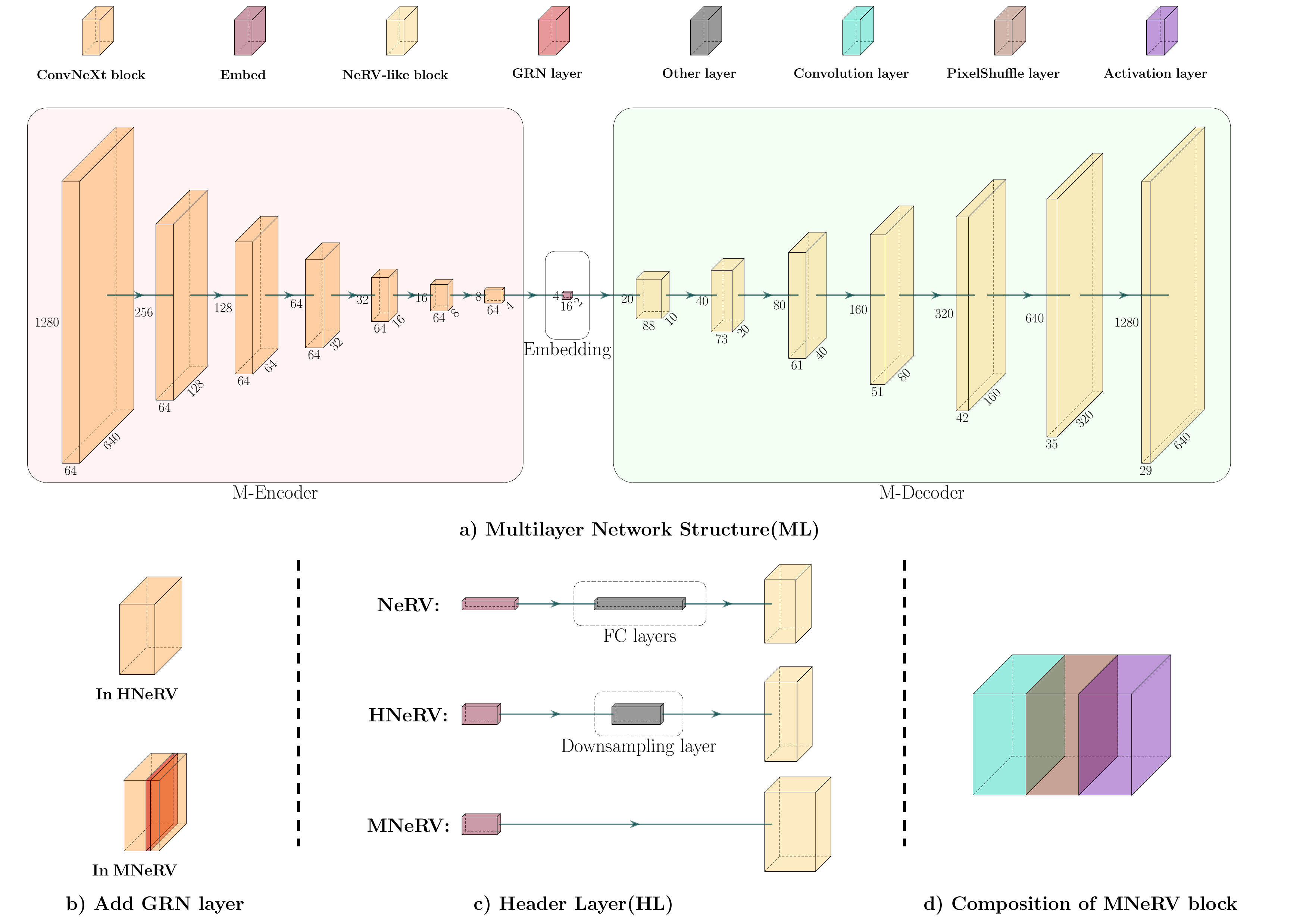}
	\caption{a) Architecture of MNeRV for 640\text{$\times$}1280. The M-Encoder consists of seven convnext blocks and the M-Decoder consists of seven MNeRV blocks. The step size of their encoding and decoding are both {5,2,2,2,2,2}. b) We introduce the GRN layer to M-Encoder. c) We demonstrate the de-redundancy design in header layer. d) We show the composition of the MNeRN block.}
	\label{fig02}
\end{figure*}

\quad First, we introduced M-Encoder, including encoding step size and encoding layer number. Then, we detailed M-Decoder and the upgraded design and parameter configuration of MNeRV blocks. The improved design of MNeRV is illustrated in Figure \ref{fig02} a), b), and c), while the composition of the MNeRV block is shown in d). Finally, we introduced the loss functions used in downstream video tasks such as video compression, video interpolation, and video restoration.
\subsection{Encoder}\label{3.1}
\quad Inspired by HNeRV, we also use ConvNeXt blocks to construct the encoding layer and assemble them into M-Encoder to extract features as the encoder end of MNeRV. In M-Encoder, to make the features extracted by the encoder more compatible with the decoder, we changed the step size {5,4,4,2,2} in HNeRV to {5,2,2,2,2,2,2}, while keeping the small embedding 16x2x4 unchanged. To enhance the competition between feature channels, we introduced the GRN layer in M-Encoder, which we proved to be necessary (Table \ref{tab05}).

\subsection{Decoder}\label{3.2}
\begin{figure}
	\centering
	\includegraphics[width=3in,height=9in,keepaspectratio]{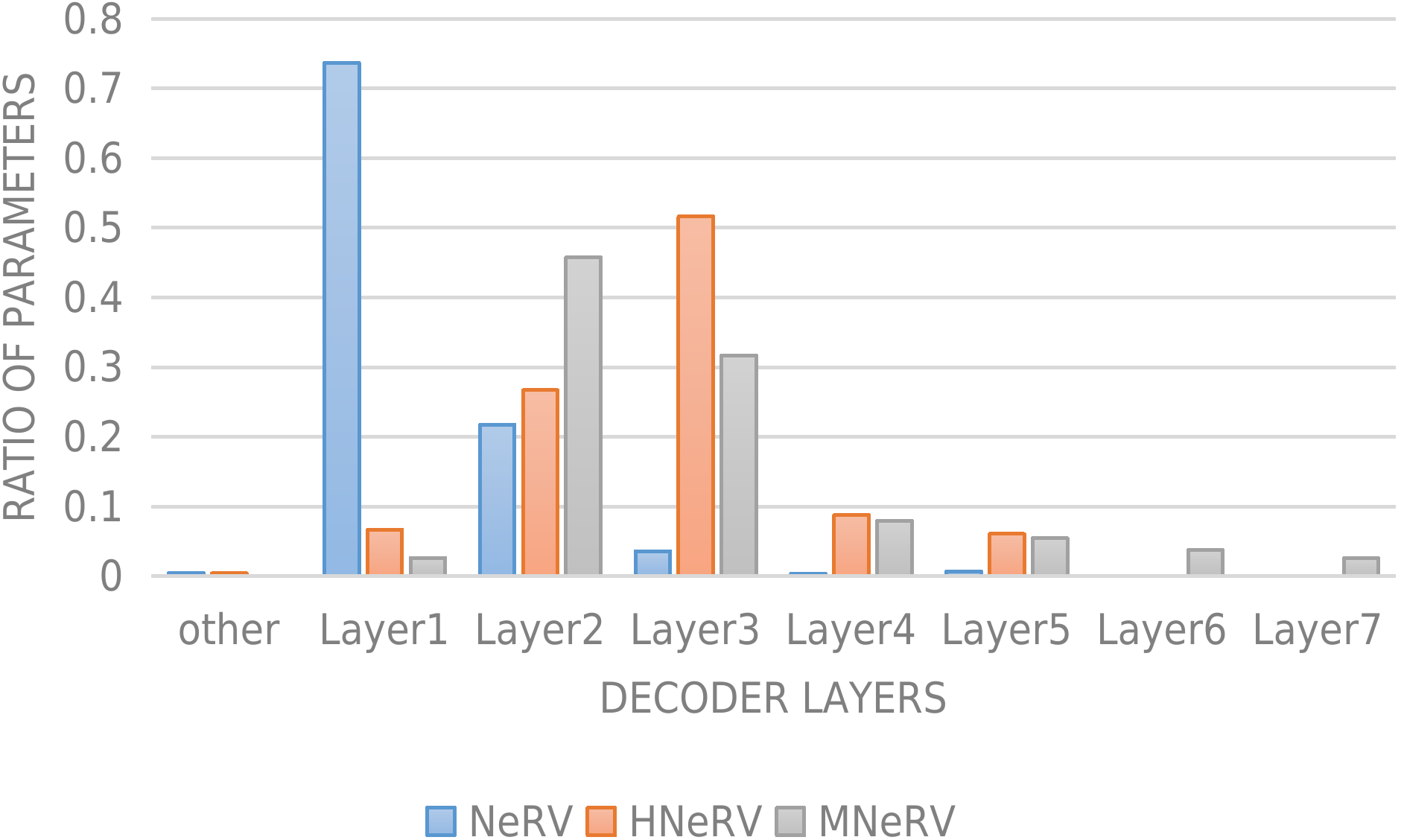}
	\caption{The parameter distribution of each layer in NeRV, HNeRV, and MNeRV. Note that in “other”, both NeRV and HNeRV perform down-sampling on frame embed, while MNeRV does not have this part.}
	\label{fig01}
\end{figure}

\quad In the NeRV decoder end, there are five decoding layers, and the kernel size k of the convolution layer in each NeRV block of each decoding layer is set to 3, and the channel attenuation factor r is set to 2, that is, the number of channels is halved for each layer. This parameter allocation makes it difficult for the later decoding layers to obtain enough parameters to fit more detailed videos. The authors of HNeRV also found this problem. In the decoder end of HNeRV, the kernel size of the convolution layer in the NeRV block of the later decoding layer was increased, and the channel attenuation factor was reduced, achieving certain results. But we proved that this is not enough. The too few decoding layers in the decoder end make it difficult to allocate parameters reasonably, and the uneven parameter allocation directly affects the fitting effect of the video. In this paper, we propose a new decoder: M-Decoder and a more efficient decoding block: MNeRV block. M-Decoder has seven decoding layers, that is, the feature map transmitted from the encoder end is enlarged seven times to obtain the fitted image. The comparison of the model parameter allocation of NeRV, HNeRV, and M-Decoder is shown in Figure \ref{fig01}. Thanks to more decoding layers and more reasonable MNeRV block size, it can use more subtle operations on images at different stages, so that the parameters are evenly distributed in each layer. Moreover, unlike the downsampling operation before the first decoding layer in HNeRV, there is no downsampling operation in M-Decoder, and all decoding layers perform upsampling operations, further reducing redundancy.

\subsection{Loss Function}\label{3.3}
Regarding the loss function, compared to $L_{2}$, which has a larger penalty for outliers and makes the model unstable, we use a combination of $L_{1}$ and $L_{MS}$ to avoid such situations. The loss function is as follows:
\begin{equation}
\begin{aligned}
&\mathrm{L}_{all}=\mathrm{L}_{1}+\mathrm{L}_{MS}\\
&\mathrm{L}_{1}=\frac{1}{\mathrm{~T}} \sum_{\mathrm{t}=1}^{\mathrm{T}} \alpha\left\|\hat{\mathrm{v}}_{\mathrm{t}}-\mathrm{v}_{\mathrm{t}}\right\|_1\\
&\mathrm{L}_{MS}=(1-\alpha)\left(1-\mathrm{MS} \_\operatorname{SSIM}\left(\hat{\mathrm{v}}_{\mathrm{t}}, \mathrm{v}_{\mathrm{t}}\right)\right)\\
&\mathrm{L}_{2}=\frac{1}{\mathrm{~T}} \sum_{\mathrm{t}=1}^{\mathrm{T}}(\hat{\mathrm{v}}_{\mathrm{t}}-\mathrm{v}_{\mathrm{t}})^2
\end{aligned}
\end{equation}

where  is the video frame fitted by the neural network,  is the corresponding ground truth. T represents the total number of frames in the video. In MNeRV, we set \text{$\alpha$} to 0.7 or 0.6. More detailed information will be explained in the experiment section.

\section{Experiments}
\subsection{Datasets and Implementation Details}
\quad We validated our model on UVG\cite{10.1145/3339825.3394937}, DAVIS\cite{7532610}, RED\cite{Nah_2019_CVPR_Workshops_REDS}, and Bunny\cite{111} datasets. The experiments were run on an RTX3090 device. In the UVG dataset, we followed the processing method of {HNeRV} and cropped 7 videos to 640\text{$\times$}1280 resolution before inputting them into the M-Encoder. We also used the Adam optimizer to train the model, set the channel attenuation factor r to 1.2, the learning rate at 0.001, and the batch size as 2. To fit 640\text{$\times$}1280 videos, we set the stride to {5,2,2,2,2,2,2} and the kernel size to {1,5,5,3,3,3,3}. In implicit neural representation of videos, the fitting speed of the neural network is the encoding speed of the video. To pursue faster encoding, we set the training rounds to 100 epochs. Unless otherwise specified, we performed the same operation on the DAVIS, RED, and bunny datasets.

\subsection{Main Results}
\quad We first conducted experiments on the bunny dataset, comparing it at 1.5M, and 3M sizes. The comparison results are shown in Table \ref{tab01}, where the loss functions of NeRV and HNeRV are L2, and the loss functions of NeRV\text{$\dagger$}, HNeRV\text{$\dagger$}, and MNeRV\text{$\dagger$} are set \text{$\alpha$} to 0.7. For fair comparison, we trained each of the three models three times with each loss function and took the average. We found that the loss function set \text{$\alpha$} to 0.7 can improve the loss function on the Bunny dataset compared to the L2 loss function. Moreover, the MNeRV model performs better than NeRV and HNeRV.

\begin{table}[]
	\caption{\centering{Performance of NeRV, HNeRV and MNeRV on the Bunny dataset. The “\text{$\dagger$}” symbol indicates the use of different loss.}}
	\begin{tabular}{c|ccccc}
		\toprule[1pt]
		
		Bunny & NeRV  & NeRV\text{$\dagger$} & HNeRV & HNeRV\text{$\dagger$} & MNeRV\text{$\dagger$} \\ 
		\cmidrule(lr){1-6}
		1.5M       & 27.86 & 29.58 & 31.06 & 31.98  & \textbf{32.14}$\uparrow$ \\ 
		3M         & 28.79 & 31.97 & 32.47 & 34.21  & \textbf{34.36}$\uparrow$	\\
		\bottomrule[1pt]
	\end{tabular}
	\label{tab01}
\end{table}

\begin{figure*}
	\centering
	\includegraphics[width=130mm,trim=0 850 0 720,clip]{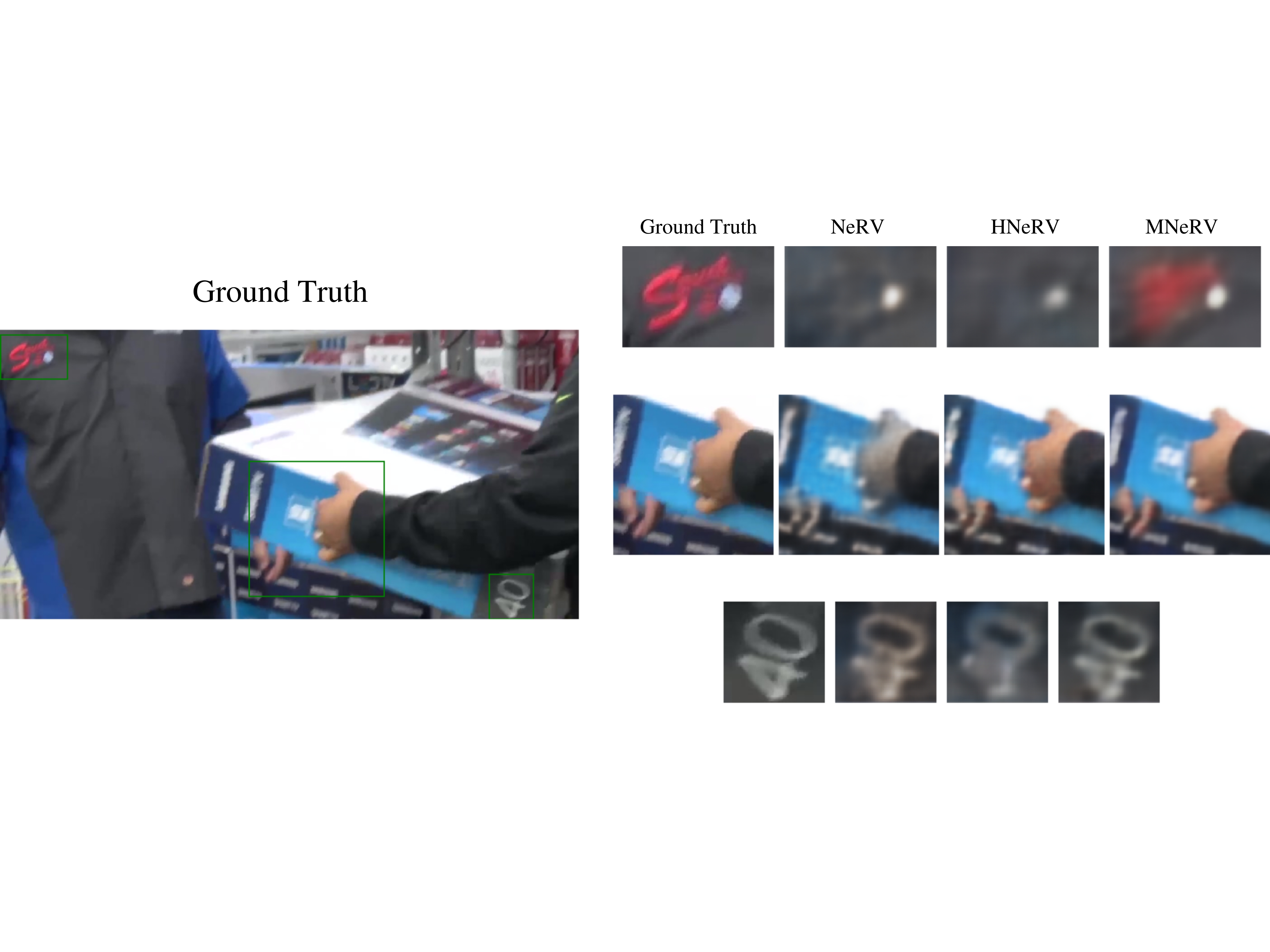}
	\caption{Visualization of video neural representations. On the left, we show the original frames. On the right, we compare NeRV, HNeRV, and MNeRV for 3 patches. }
	\label{fig03}
\end{figure*}

\begin{table*}[]
	\centering
	\caption{\centering{Comparison results of 10 sub-datasets of REDs.}}
	\begin{tabular}{c|c|cccccccccc|c}
		\toprule[1pt]
		& size  & 032   & 033   & 084   & 108   & 153   & 231   & 232   & 233   & 243   & 269   & avg   \\
		\cmidrule(lr){1-13}
		HNeRV      & 1.50M & 23.49 & 25.24 & 20.64 & 25.52 & 21.1  & 26.1  & 24.91 & 25.48 & 21.98 & 24.71 & 23.92 \\
		MNeRV      & \textbf{1.46M} & \textbf{25.9}  & \textbf{27.3}  & \textbf{22.67} & \textbf{28}    & \textbf{23.41} & \textbf{28.38} & \textbf{27.02} & \textbf{27.61} & \textbf{23.99} & \textbf{26.62} & \textbf{26.09} \\ 
		\bottomrule[1pt]
	\end{tabular}
	\label{tab02}
\end{table*}
\begin{table*}[]
	\centering
	\caption{\centering{Comparison results of UVG datasets.}}
	\begin{tabular}{c|c|ccccccc|c}
		\toprule[1pt]
		& size  & beauty & Bosphorus & HoneyBee & Jockey & ReadySteadyGo & ShakeNDry & YachtRide & avg   \\
		\cmidrule(lr){1-10}
		HNeRV & 1.50M & 33.92  & 32.49     & \textbf{38.13}    & 28.75  & 23.82         & \textbf{33.01}     & 28.45     & 31.22 \\
		MNeRV & \textbf{1.48M} & \textbf{34.1}   & \textbf{32.79}     & 37.79    & \textbf{30.6}   & \textbf{24.75}         & 32.98     & \textbf{28.84}     & \textbf{31.69} \\ 
		\bottomrule[1pt]
	\end{tabular}
	\label{tab03}
\end{table*}

\begin{figure}[]
	\centering
	\includegraphics[width=60mm]{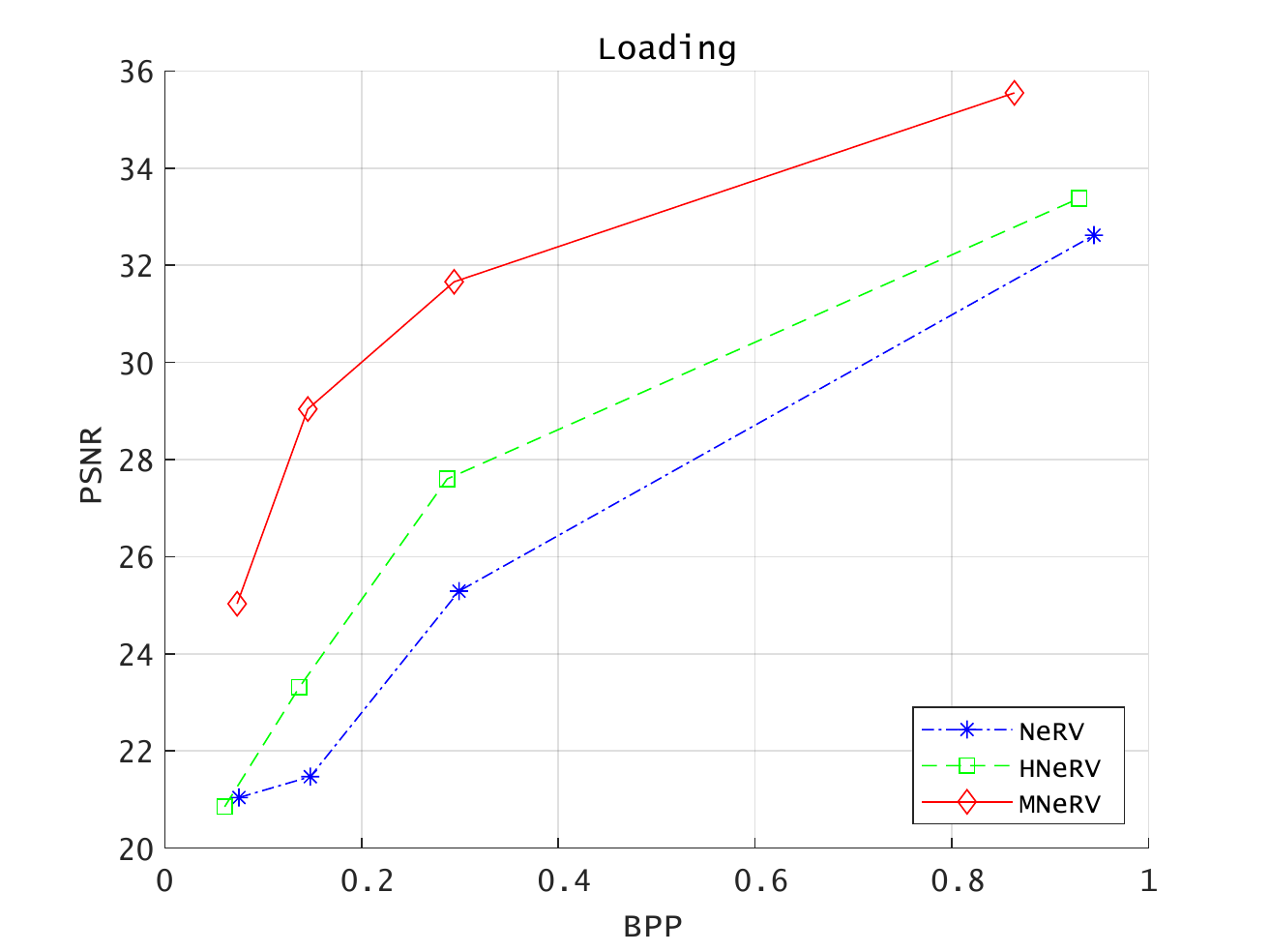}
	\caption{Compression results on loading dataset}
	\label{fig04}
\end{figure}

We conducted comparative experiments on four scales of the Loading dataset. For the loss function part, we set \text{$\alpha$} to 0.6. Figure \ref{fig04} shows the performance of NeRV, HNeRV, and MNeRV models on different scales, and their visual comparison is shown in Figure \ref{fig03}. The experimental results on the Loading dataset show that our method has a significant improvement over NeRV and HNeRV. In addition, we compared our method with HNeRV on 10 REDs sub-datasets and 7 UVG sub-datasets, using L2 as the loss function, and the results are shown in Table \ref{tab02} and Table \ref{tab03}, respectively. Note that our method achieves better results with fewer parameters. The model size is indicated in the second column of both tables. It should be noted that the ShakeNDry dataset is different from the other UVG sub-datasets, as it only contains 300 images. On the ShakeNDry dataset, the size of HNeRV is 1.52M, and the size of MNeRV is 1.49M.

\subsection{Ablation Studies}
\quad We conducted detailed ablation analysis comparative experiments on NeRV, HNeRV, and MNeRV. In implicit neural networks, the main parameters for adjusting parameter allocation are kernel size and stride. Thanks to the design of its multi-layer network structure(ML), MNeRV offers a more extensive range of parameter adjustment options. We conducted ablation on kernel size and stride on the jockey dataset, and the results are shown in Table \ref{tab04}. We found that in the 5-layer NeRV-like block structure, due to the natural limitation of the number of layers, the parameter allocation scheme is lacking, and it is easy to have bloated parameters (as shown in Figure \ref{fig01}). The 7-layer NeRV-like block structure can adjust more kernel sizes and strides, and there are more parameter allocation methods, and the parameter volume of each layer is more uniform, resulting in better effects. Finally, we selected {5,2,2,2,2,2,2} and {1,5,5,3,3,3,3} as the stride and kernel size of MNeRV, respectively. 

\begin{table}[]
	\centering
	\caption{\centering{Ablation experiments on stride and kernel size.}}
	\begin{tabular}{cccc}
		\toprule[1pt]
		block & s             & K\_s          & PSNR  \\
		\cmidrule(lr){1-4}
		\multicolumn{1}{c}{\multirow{4}{*}{5layers}}
		& 5,4,2,2,2,2   & 1,3,5,5,5,5   & 28.64 \\
		& 5,4,2,2,2,2   & 1,3,5,7,5,3   & 28.35 \\
		& 5,4,2,2,2,2   & 1,3,5,5,3,3   & 28.77 \\
		& 5,4,2,2,2,2   & 1,3,5,5,5,3   & 28.67 \\
		\cmidrule(lr){1-4}
		\multirow{7}{*}{7layers}                     
		& 5,2,2,2,2,2,2 & 1,3,5,5,5,5,5 & 29.23 \\
		& 5,2,2,2,2,2,2 & 1,3,5,5,5,3,3 & 29.2  \\
		& 5,2,2,2,2,2,2 & 1,3,5,5,3,3,3 & 29.23 \\
		& 5,2,2,2,2,2,2 & 1,3,5,7,5,3,3 & 28.9  \\
		& 5,2,2,2,2,2,2 & 1,3,5,3,3,3,3 & 29.3  \\
		& 5,2,2,2,2,2,2 & 1,3,3,3,3,3,3 & 29.11 \\
		& 5,2,2,2,2,2,2 & 1,5,5,3,3,3,3 & \textbf{29.62} \\
		\bottomrule[1pt]
	\end{tabular}
	\label{tab04}
\end{table}
The effects of three novel components of our proposed method are studied separately: the GRN layer introduced in M-Encoder, the multi-layer network structure design and the header layer(HL) design at the beginning of the network. The ablation results are shown in Table \ref{tab02}, which shows the lifting effect of each component in detail.

To evaluate the performance of MNeRV on videos with significant camera motion, we conducted a comparative experiment on the REDs dataset (Figure \ref{fig05}) and performed an ablation study on the loss function (Table \ref{tab06}).

\begin{figure*}[ht]
	\centering
	\includegraphics[width=155mm,trim=0 0 420 0,clip]{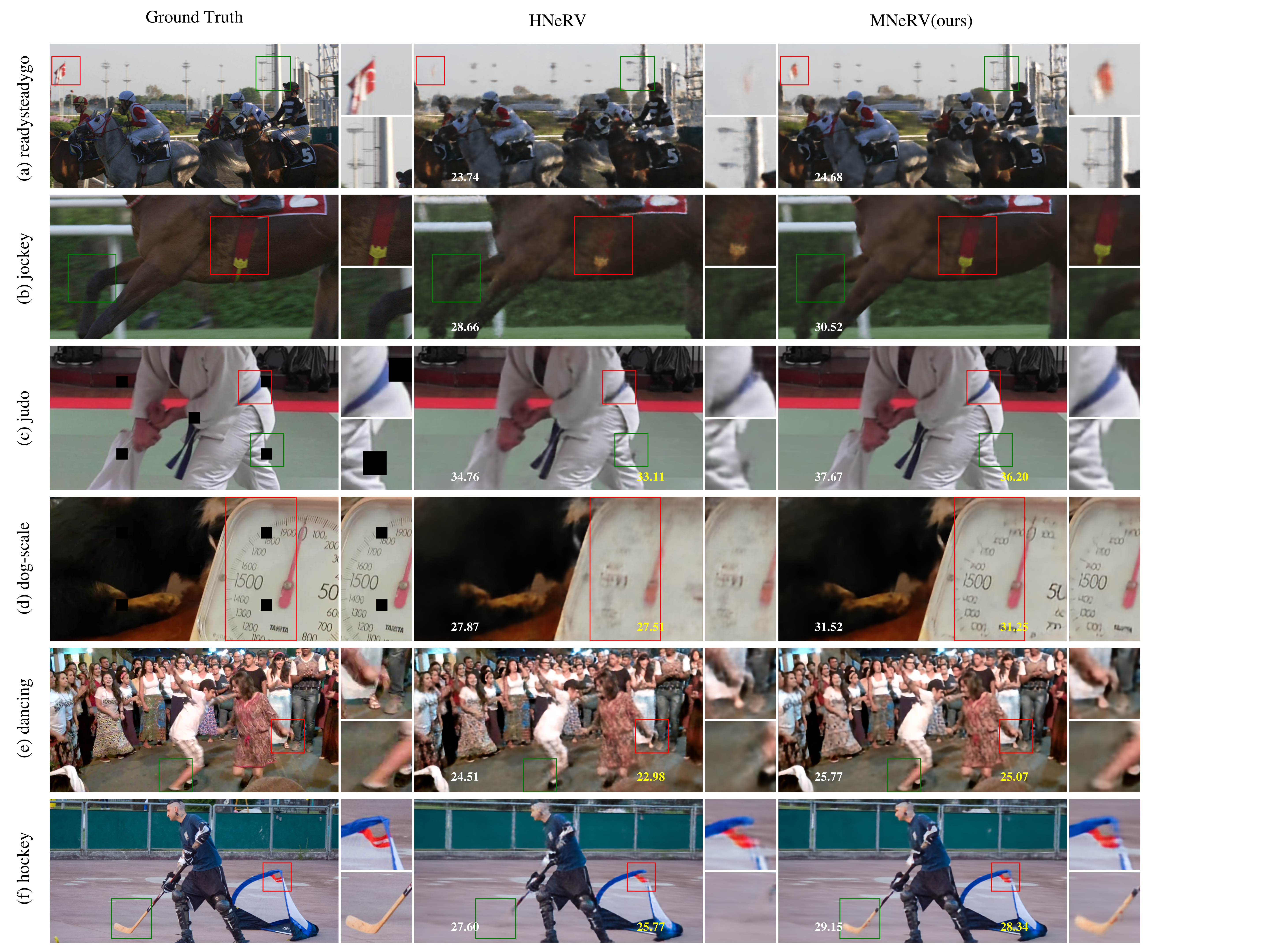}
	\caption{A visual comparison of the results of video compression (a, b), video restoration (c, d), and interpolation (e, f). Left) ground truth.Middle) HNeRV output.Right) MNeRV output. The white numbers indicate the best PSNR of each method trained on the original video, and the yellow numbers indicate the results after restoration or interpolation.}
	\label{fig08}
\end{figure*}
\begin{table}[]
	\centering
	\caption{\centering{Ablation study on components.}}
	\begin{tabular}{c|ccccc}
		\toprule[1pt]
		UVG       & GRN & ML & HL & PSNR  & MSSSIM     \\
		\cmidrule(lr){1-6}
		HNeRV     & \XSolidBrush   & \XSolidBrush  & \XSolidBrush  & 32.41 & 0.9085143  \\
		Variant 1 & \Checkmark   & \XSolidBrush  & \XSolidBrush  & 32.45 & 0.90927143 \\
		Variant 2 & \Checkmark   & \Checkmark  & \XSolidBrush  & 32.61 & 0.91531    \\
		MNeRV     & \Checkmark   & \Checkmark  & \Checkmark  & \textbf{32.64} & \textbf{0.915486}   \\ 
		\bottomrule[1pt]
	\end{tabular}
	\label{tab05}
\end{table}

\begin{figure}[]
	\centering
	\includegraphics[width=60mm]{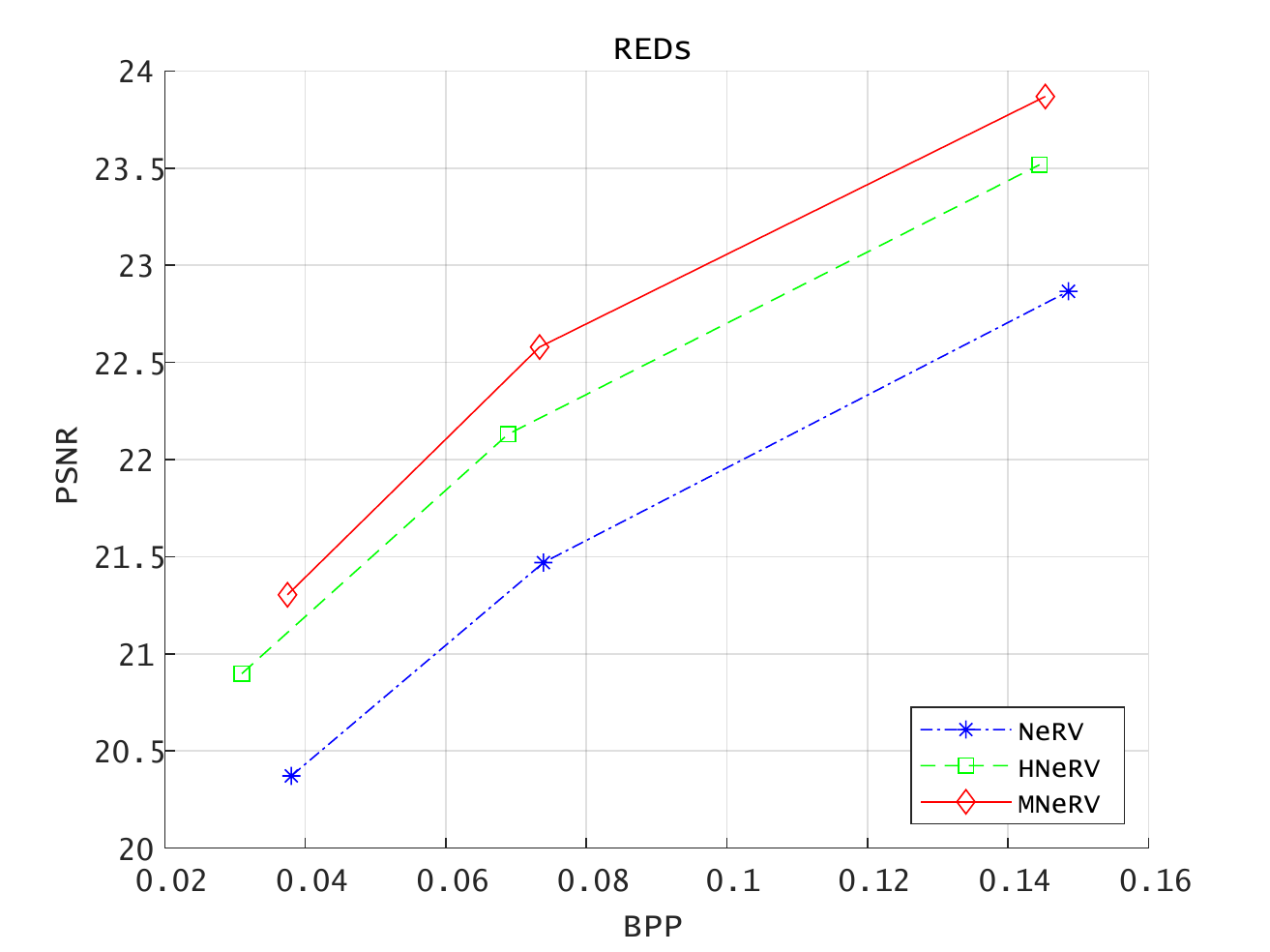}
	\caption{Compression results on REDs dataset.}
	\label{fig05}
\end{figure}

\begin{table}[]
	\centering
	\caption{\centering{Detailed ablation experiments on the loss function are carried out on the REDs dataset, where sml1 represents the smooth L1 function,s represents the SSIM function, and ms represents the MSSSIM function.}}
	\begin{tabular}{cc|cc}
		\toprule[1pt]
		Loss & PSNR & Loss & PSNR  \\
		\cmidrule(lr){1-4}
		0.7*L1+0.3*SSIM & 22.875  & 0.5*L2+0.5*S  & 21.752 \\
		SML1    & 23.191 & 0.3*L1+0.7*S  & 21.602\\
		L1      & 22.565 & 0.3*L2+0.7*S  & 21.286 \\
		0.5*L1+0.5*S  & 22.408 &   0.6*L2+0.4*MS       & 23.503  \\
		0.7*L2+0.3*S  & 22.706 & 0.9*L1+0.1*MS  & 23.319    \\
		0.7*L2+0.3*L1 & 23.039 & 0.8*L1+0.2*MS       & 23.502 \\
		0.5*L2+0.5*L1 & 22.92  & 0.4*L2+0.6*MS       & 22.853 \\
		0.9*L1+0.1*S   & 22.842 & 0.7*L1+0.3*MS & \textbf{23.729} \\
		
		\bottomrule[1pt] 
	\end{tabular}
	\label{tab06}
\end{table}

\subsection{Downstream Tasks}
\quad We compared MNeRV with other implicit neural representation models on various downstream tasks, including video interpolation, video restoration, and video compression.

\begin{figure}[h]
	\centering
	\includegraphics[width=60mm]{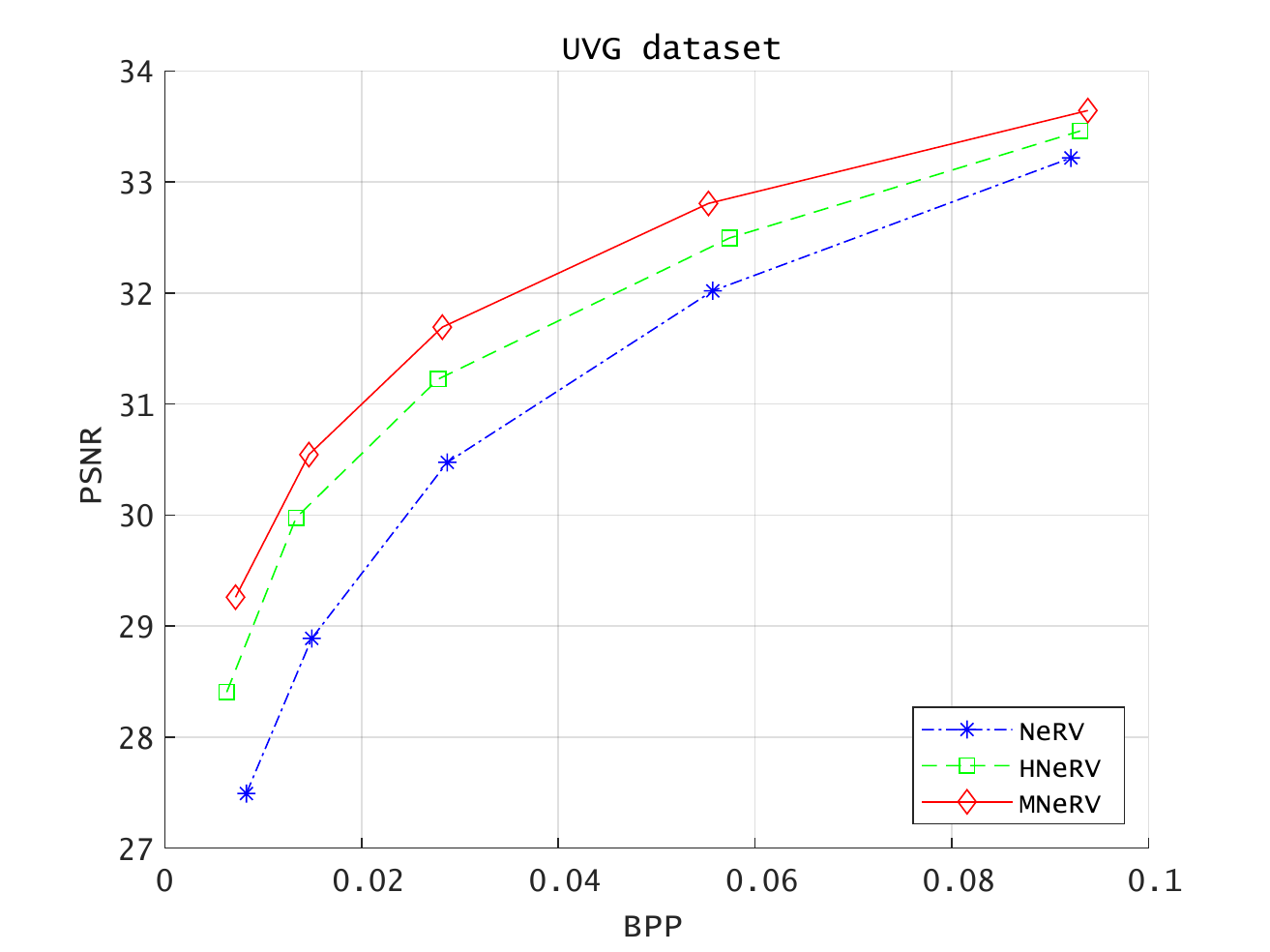}
	\includegraphics[width=60mm]{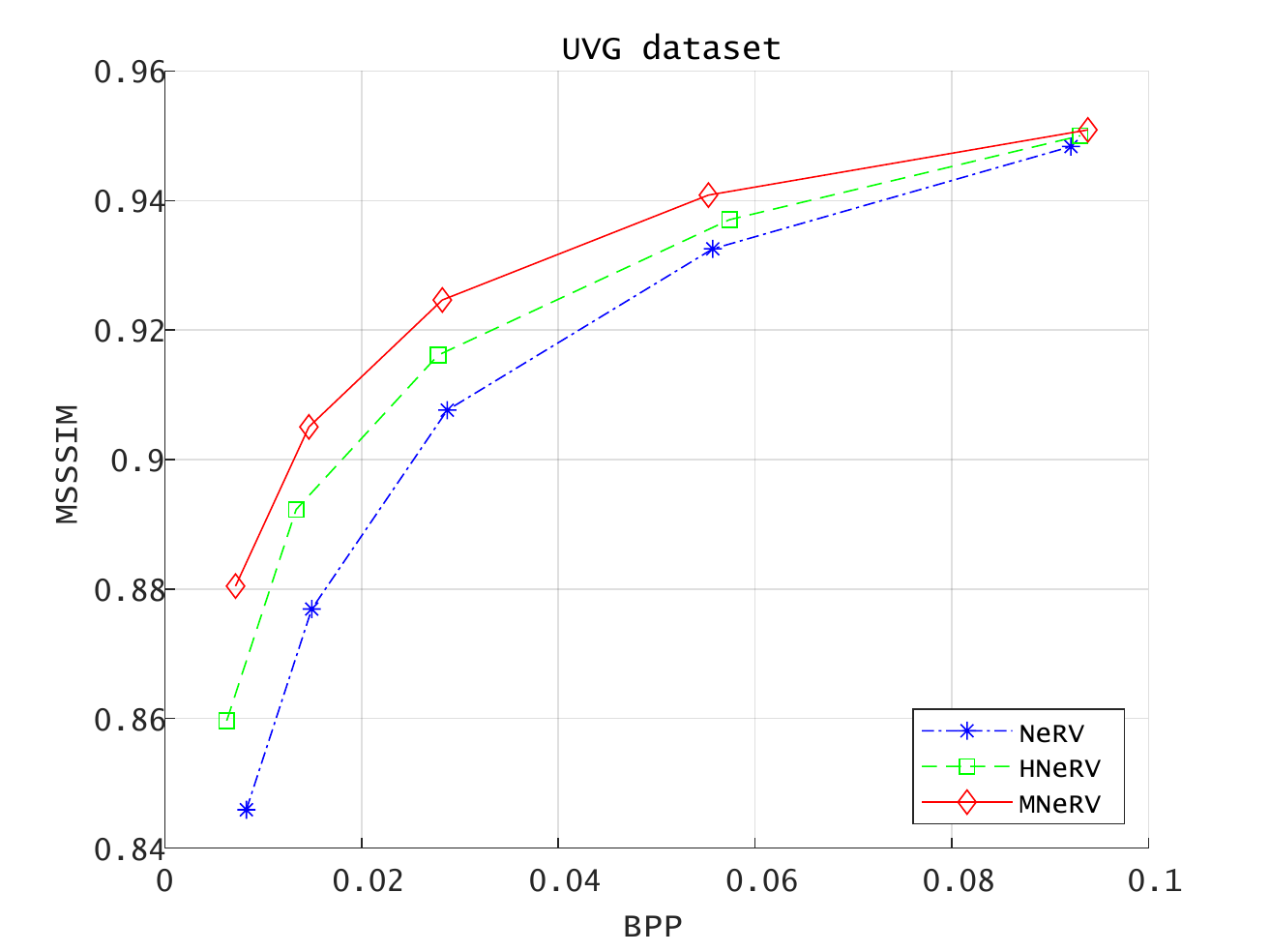}
	\caption{Compression results on UVG dataset.}
	\label{fig06}
\end{figure}

\begin{figure}[h]
	\centering
	\includegraphics[width=60mm]{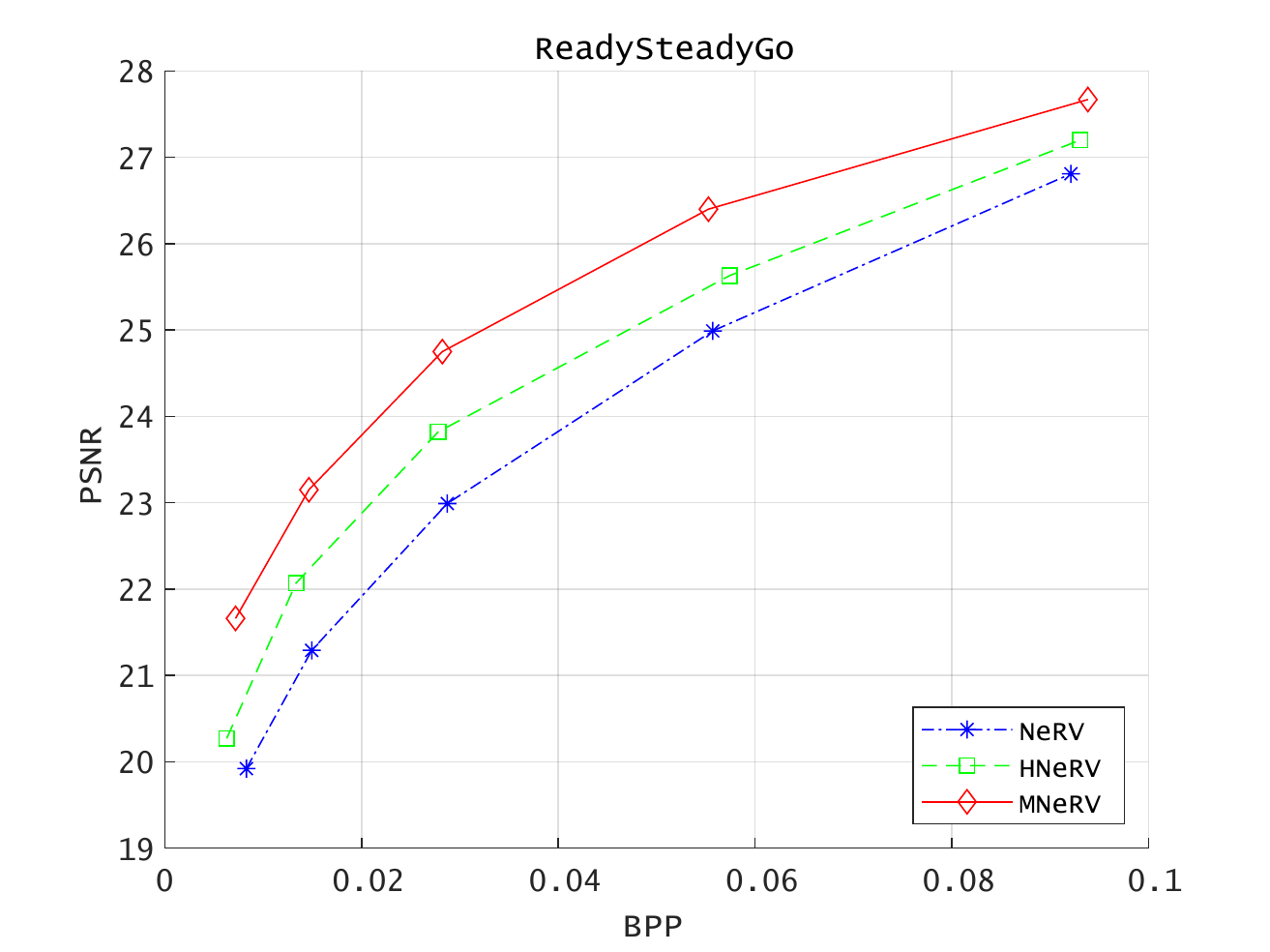}
	\includegraphics[width=60mm]{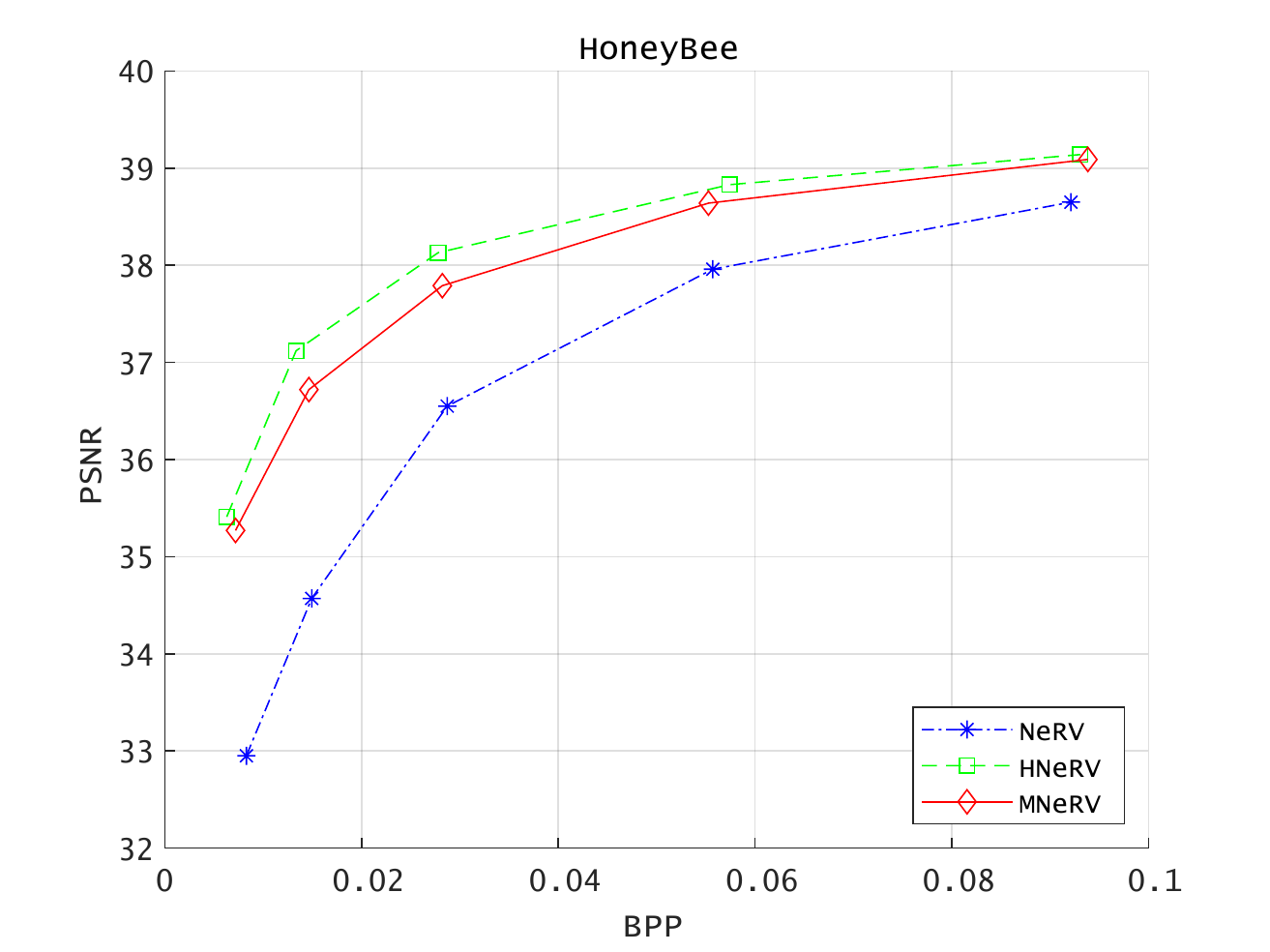}
	\caption{Best/worst compression cases from UVG dataset.}
	\label{fig07}
\end{figure}

\textbf{video compression.} Figure \ref{fig08}(a,b) shows the comparison of the reconstruction quality of HNeRV and MNeRV. In the video compression experiment, we followed the pipeline of HNeRV and conducted a comparison experiment of NeRV, HNeRV, and MNeRV on the UVG dataset using different compression ratios. The results are shown in Figure \ref{fig06}, where our method is better than NeRV and HNeRV in both PSNR and MSSSIM metrics. In addition, we show the best and worst results of MNeRV in Figure \ref{fig07}. In videos with intense camera movement (such as the readysteadygo dataset), MNeRV has shown significant improvement over HNeRV, but in videos with less camera movement (such as the honeybee dataset), MNeRV is inferior to HNeRV. On the UVG dataset, MNeRV’s video compression performance is generally better than HNeRV’s (+0.47 PSNR).

\textbf{video restoration.}In Figure \ref{fig08}(c,d), we show the comparison of the restoration effect of HNeRV and MNeRV. MNeRV also has excellent performance.

\textbf{video interpolation.} As a type of neural representation, MNeRV has powerful frame fitting ability. In the video interpolation experiment, we used HNeRV and MNeRV as test frames every other frame, and used interpolated embedding as input. Through learnable content embedding, MNeRV showed better results, as shown in Figure \ref{fig08}(e,f).

\section{Conclusion}
\quad In this paper, we propose a multilayer neural representation for videos that utilizes a novel encoder-decoder architecture: M-Encoder and M-Decoder.  This architecture allows for a more reasonable parameter allocation, while retaining the quality of the fitted images and reducing the number of model parameters. It uses a network with more layers to fit videos, resulting in better fitting effects in some videos with camera movement. We have demonstrated through experiments that MNeRV has significantly improved on UVG, RED, DAVIS, and other datasets. Finally, we apply MNeRV to downstream tasks such as video interpolation and video restoration, and showcase its excellent performance.

Future work. The potential of the encoder end of MHeRV has not been fully realized, and future work should focus on improving the encoder end to make the information contained in each frame’s embedding more accurate.

\section*{Acknowledgment}
This project was partially funded by: “AI neural network algorithm gives optical camera laser capability from 2021 Guangdong Province Key Construction Discipline Scientific Research Ability Improvement Project”, “Neutrino detector intelligent monitoring from National key research and development program”, “Knowledge graph construction for big data of scientific and technological resources from Wuyi University High-level Talent Scientific Research Start-up Project”.

\bibliographystyle{IEEEtran}
\bibliography{MNeRV}
\clearpage
\section{Supplementary Material}

\subsection{Supplementary Material}
\quad We conducted a detailed ablation study on the variants of the three components(GRN, ML, and HL) on the UVG dataset. Tables X and A present the performance of each variant on the UVG sub-datasets.
\begin{table*}[h]
	\centering
	\caption{\centering{More detailed ablation experiments (PSNR) for each video in the UVG dataset.}}
	\begin{tabular}{c|ccccccc|c}
		\toprule[1pt]
		UVG-PSNR  & beauty & Bosphorus & HoneyBee & Jockey & ReadySteadyGo & ShakeNDry & YachtRide & avg    \\
		\cmidrule(lr){1-9}
		HNeRV     & 34.27  & 33.82     & 38.97    & 30.97  & 25.165        & 33.905    & 29.8      & 32.414 \\
		Variant 1 & 34.28  & 33.815    & 38.95    & 31.08  & 25.23         & 33.92     & 29.845    & 32.446 \\
		Variant 2 & 34.3   & 33.73     & 38.46    & \textbf{32.15}  & \textbf{26.21}         & 33.64     & 29.79     & 32.611 \\
		MNeRV     & \textbf{34.3}   & \textbf{33.86}     & \textbf{38.48}    & 32.11  & 26.11         & \textbf{33.69}     & \textbf{29.96}     & \textbf{32.644} \\
		\bottomrule[1pt]
		
	\end{tabular}
	\label{tab07}
\end{table*}

\begin{table*}[h]
	\centering
	\caption{\centering{More detailed ablation experiments (MSSSIM) for each video in the UVG dataset.}}
	\begin{tabular}{c|ccccccc|c}
		\toprule[1pt]
		UVG-MSSSIM & beauty & Bosphorus & HoneyBee & Jockey & ReadySteadyGo & ShakeNDry & YachtRide & avg        \\
		\cmidrule(lr){1-9}
		HNeRV      & 0.9076 & 0.9363    & 0.9844   & 0.8742 & 0.8391        & 0.9277    & 0.8903    & 0.9085143  \\
		Variant 1  & 0.9077 & 0.9364    & 0.9844   & 0.8757 & 0.8415        & 0.9282    & 0.891     & 0.90927143 \\
		Variant 2  & \textbf{0.9080}  & 0.9364    & 0.9830    & \textbf{0.8953} & \textbf{0.8701}        & 0.9236    & 0.8908    & 0.91531    \\
		MNeRV      & 0.9078 & \textbf{0.9385}    & \textbf{0.9831}   & 0.8944 & 0.8662        & \textbf{0.9245}    & \textbf{0.8939}    & \textbf{0.915486}   \\ 
		\bottomrule[1pt]
	\end{tabular}
	\label{tab08}
\end{table*}

\subsection{More Visualizations}
\begin{figure*}[h]
	\centering
	\includegraphics[width=160mm,trim=0 2050 210 0,clip]{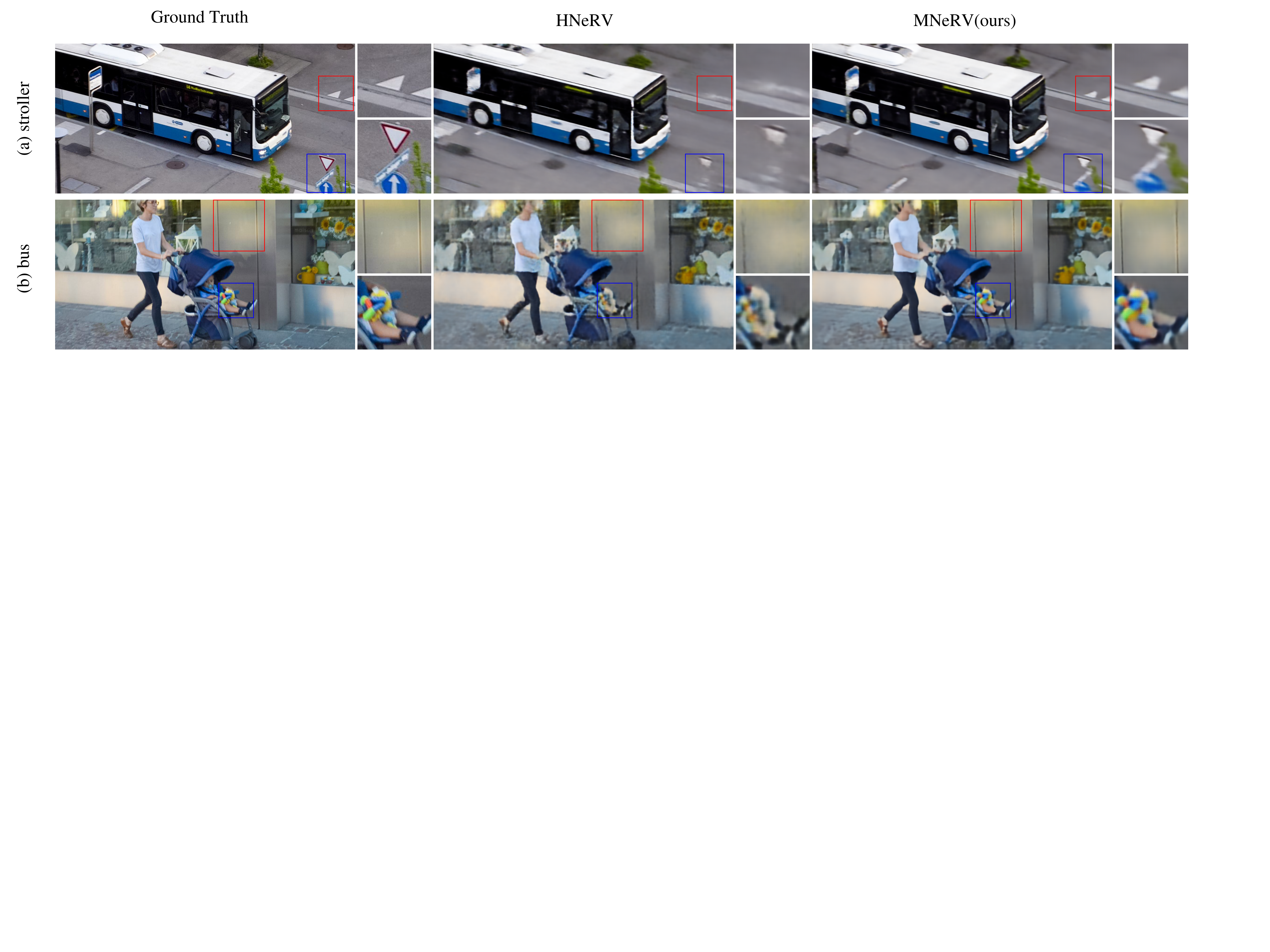}
	\caption{Video interpolation results.}
	\label{fig10}
\end{figure*}

\begin{figure*}[h]
	\centering
	\includegraphics[width=160mm,trim=0 2580 210 0,clip]{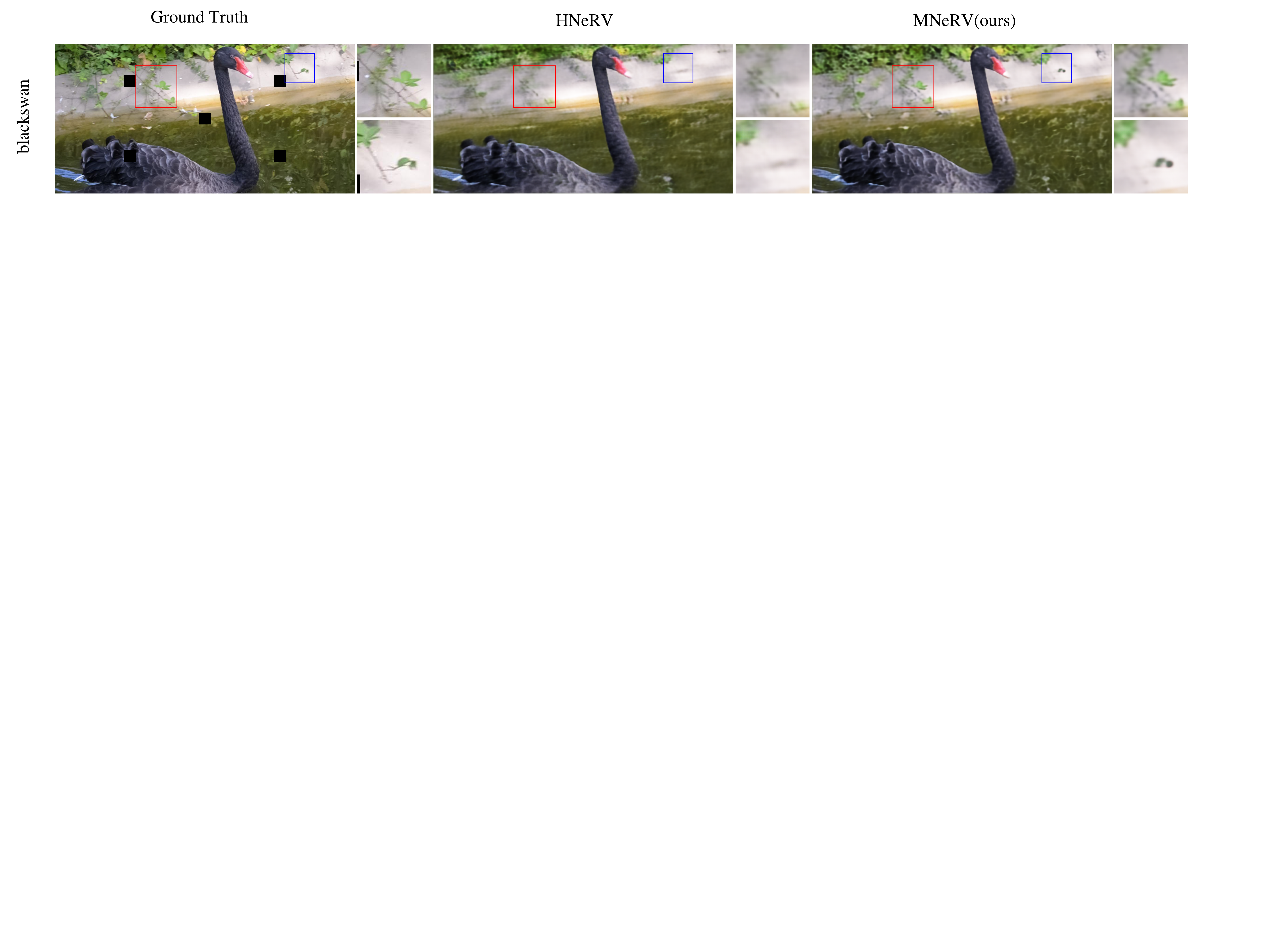}
	\caption{ Video inpainting results.}
	\label{fig11}
\end{figure*}

We show more visualizations for video interpolation (Figure \ref{fig10}), and video inpainting (Figure \ref{fig11}).

\subsection{Detailed Comparative Experiments}
\begin{figure*}
	\centering
	\includegraphics[width=70mm]{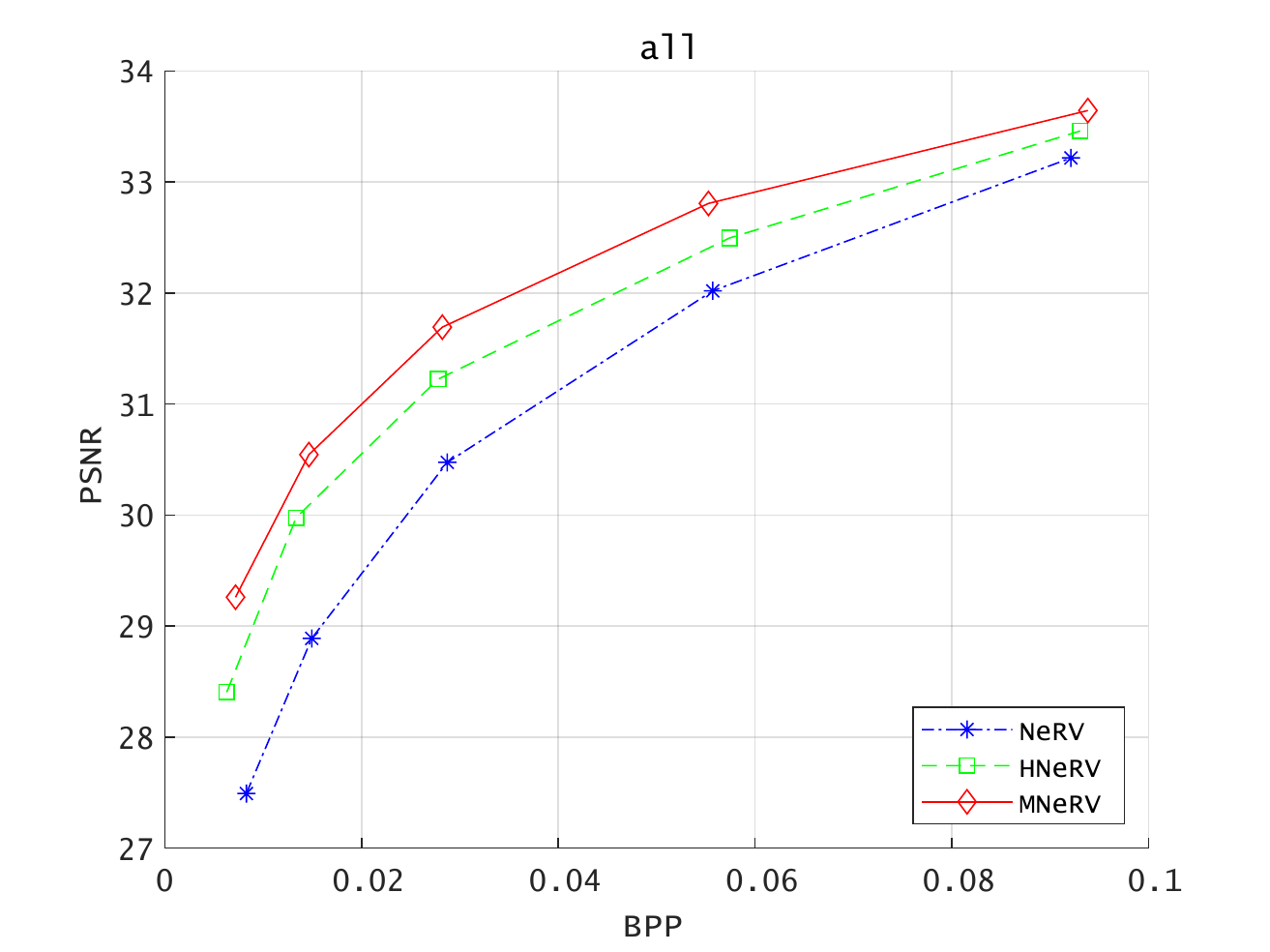}
	\includegraphics[width=70mm]{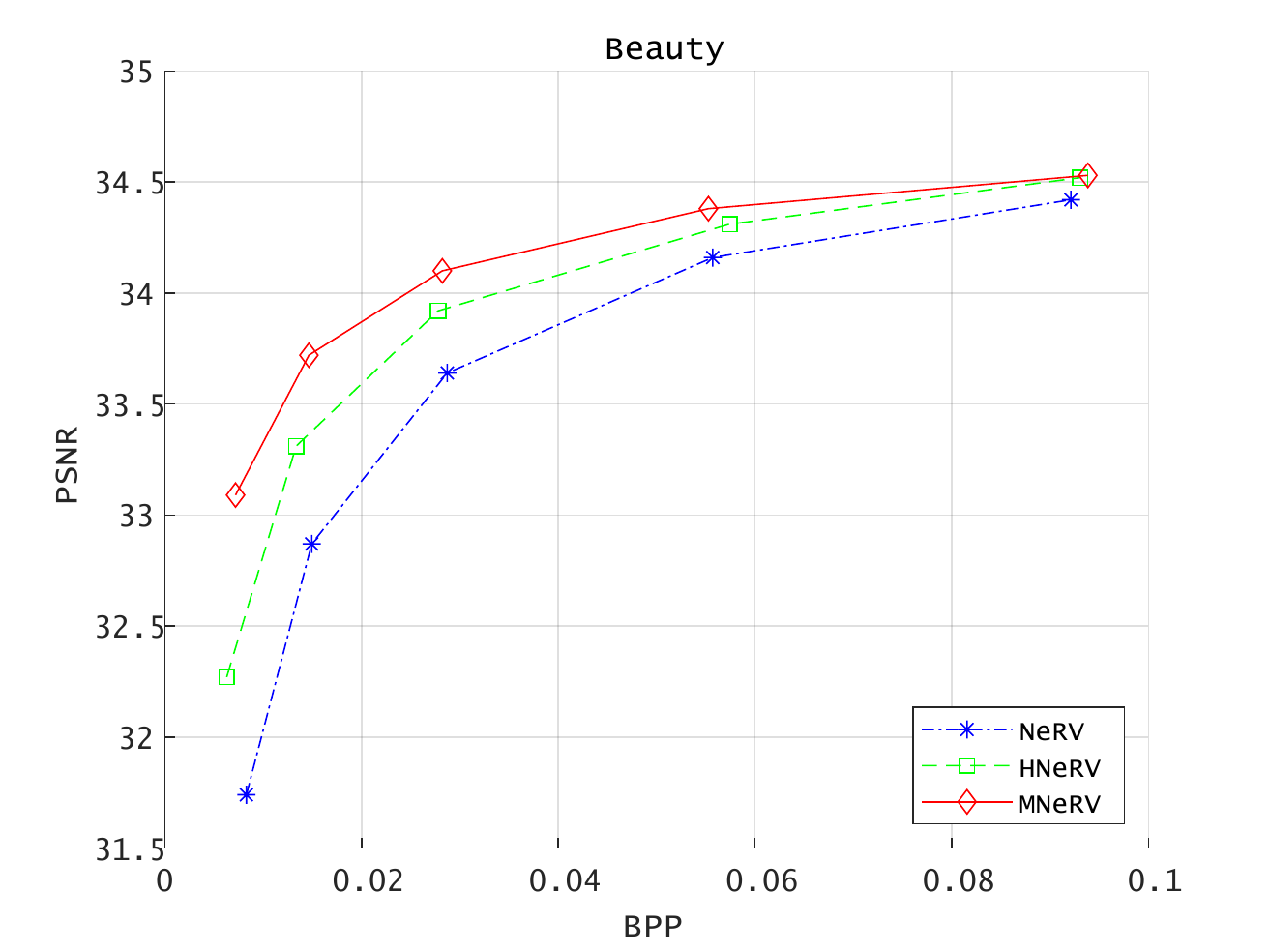}
	\includegraphics[width=70mm]{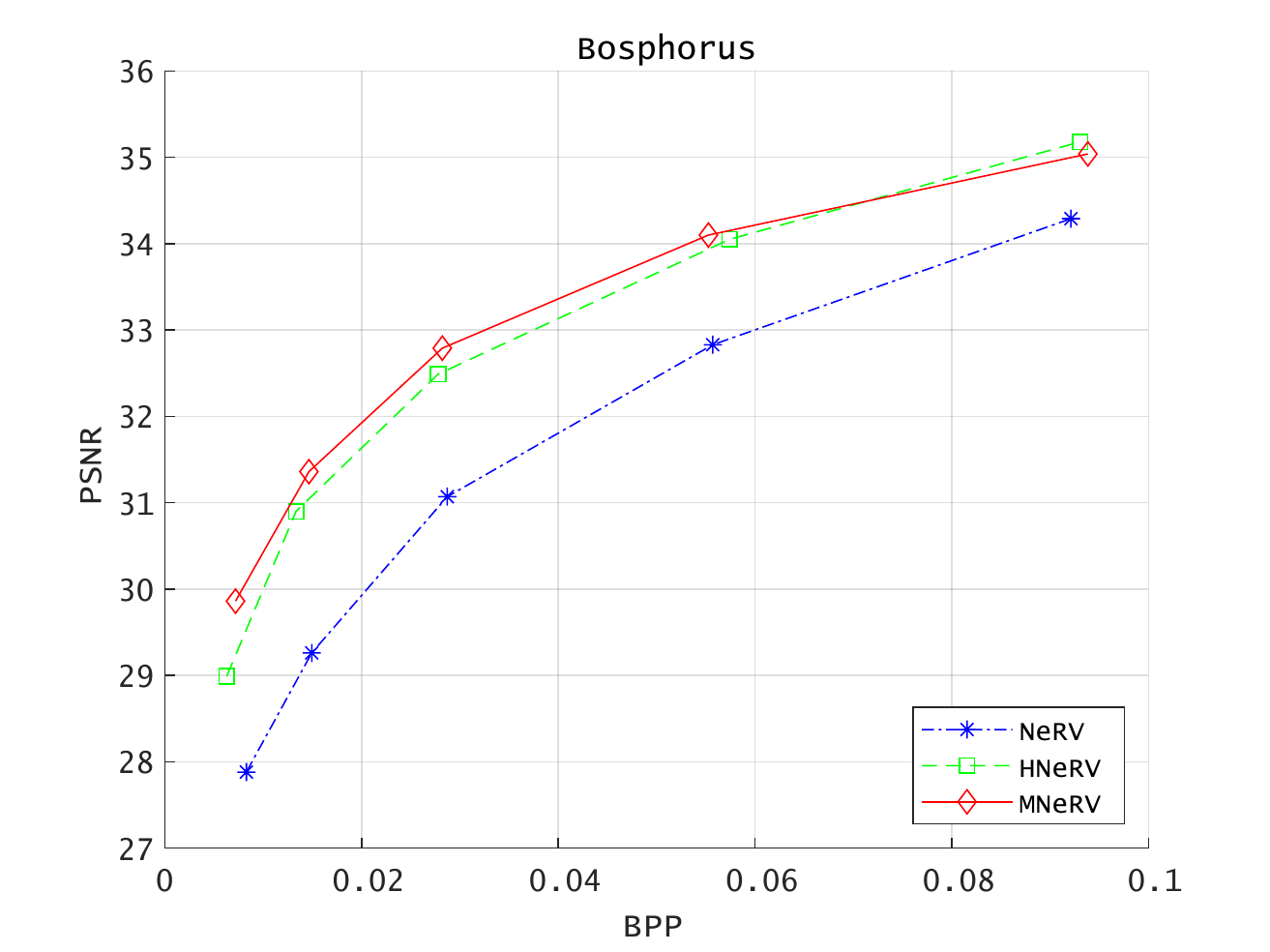}
	\includegraphics[width=70mm]{9.pdf}
	\includegraphics[width=70mm]{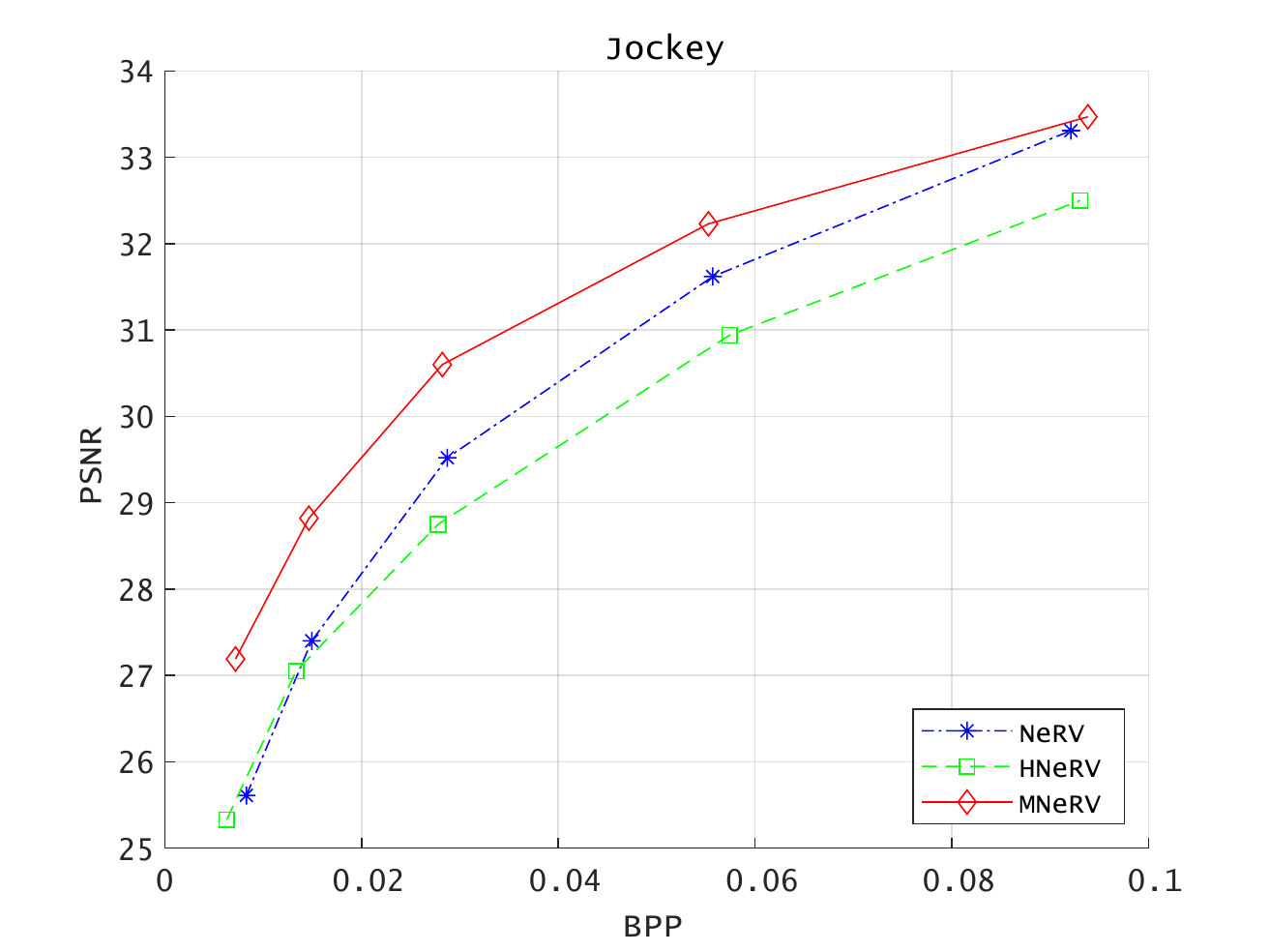}
	\includegraphics[width=70mm]{8.pdf}
	\includegraphics[width=70mm]{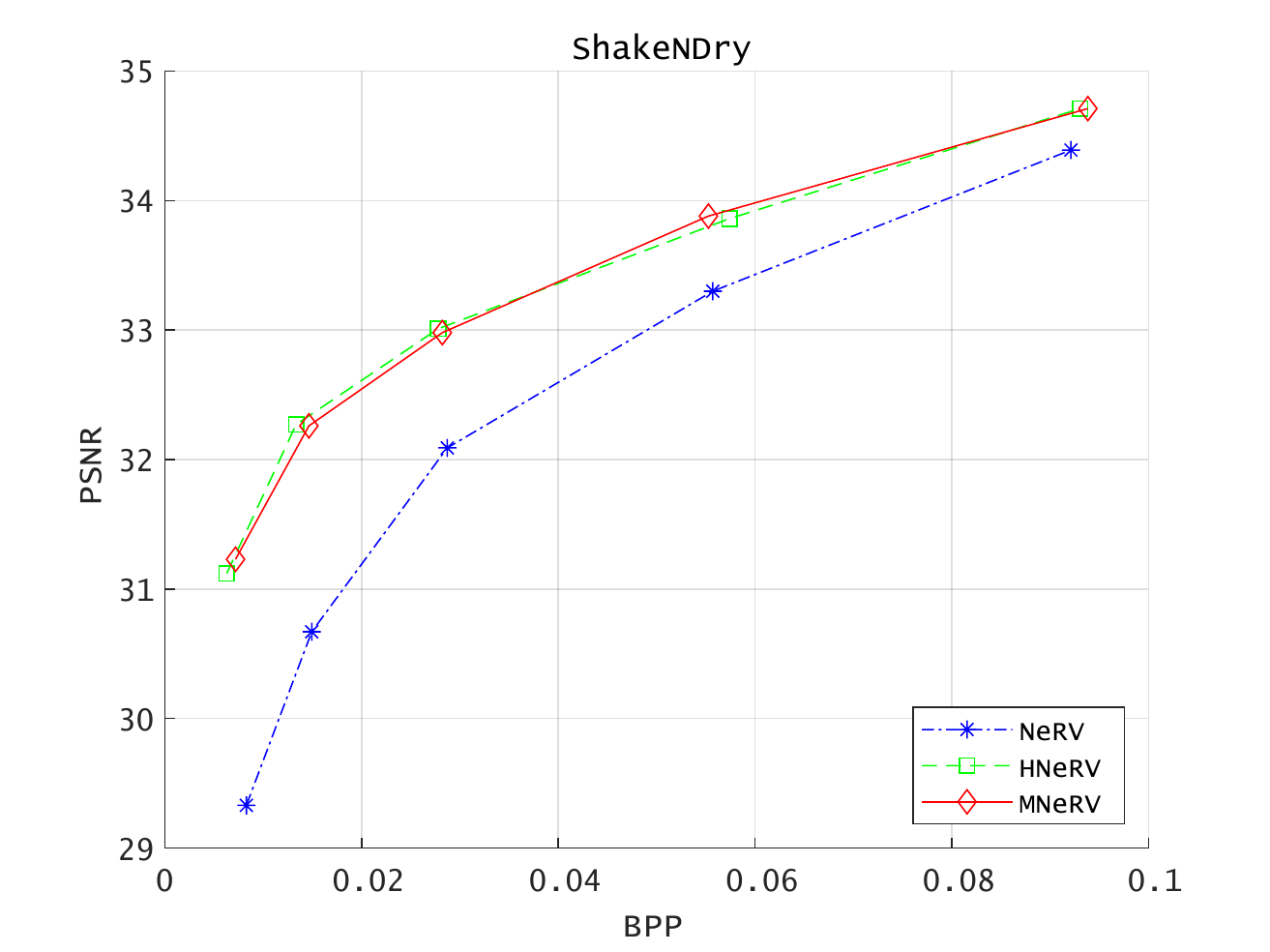}
	\includegraphics[width=70mm]{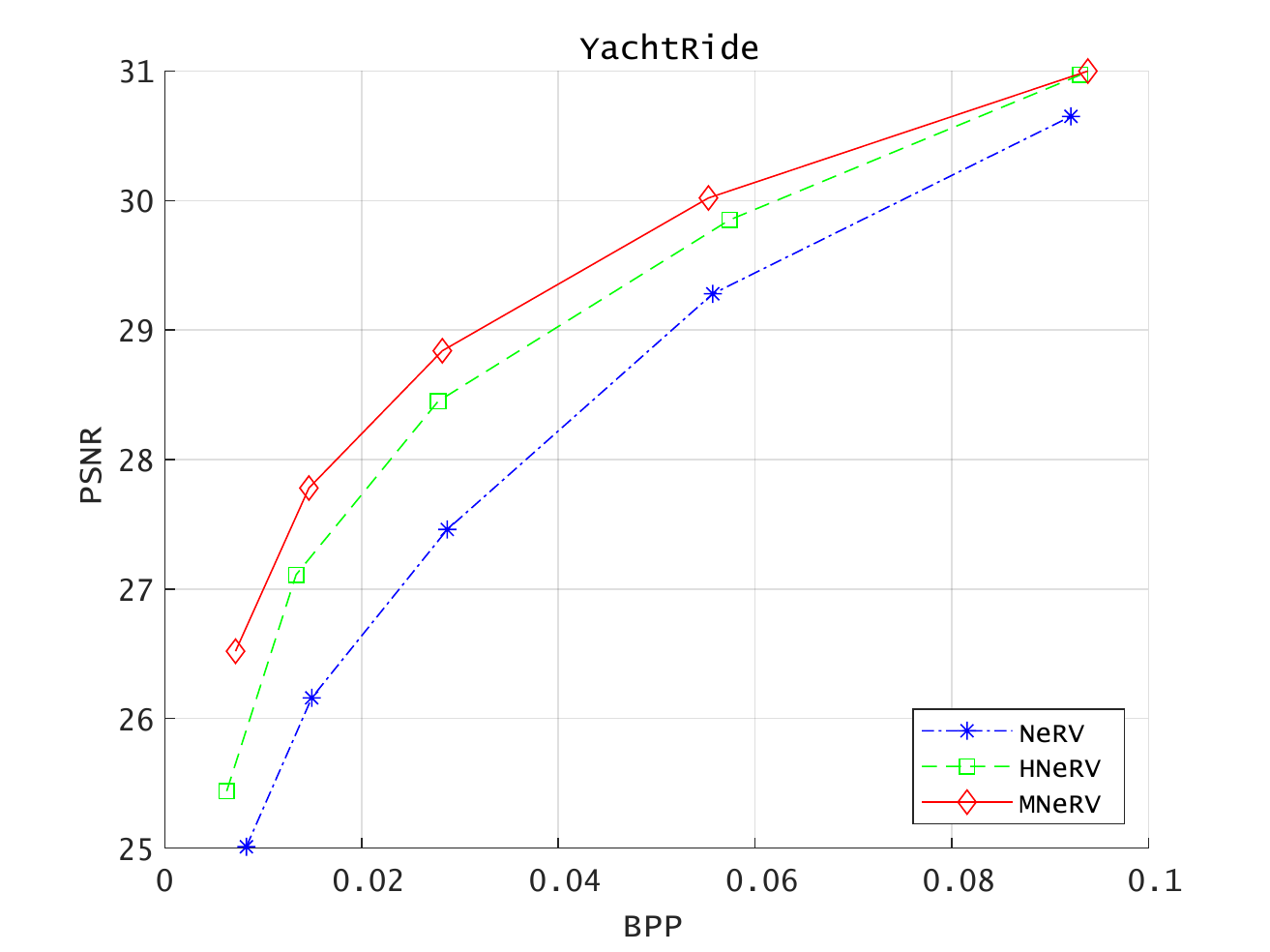}
	\caption{\centering{Video compression results for the UVG dataset and its sub-datasets.}}
	\label{fig09}
\end{figure*}

In Figure \ref{fig09}, we present the video compression results obtained on the UVG dataset(all) and each of its sub-datasets.

\end{document}